\documentclass[conference]{IEEEtran}
\IEEEoverridecommandlockouts
\usepackage{cite}
\usepackage{amsmath,amssymb,amsfonts}
\usepackage{algorithmic}
\usepackage{graphicx}
\usepackage{textcomp}
\usepackage{xcolor}

\usepackage{bm}
\usepackage{url}
\usepackage{breakurl}
\usepackage[breaklinks]{hyperref}
\usepackage{float}
\usepackage{stfloats} 
\usepackage{makecell}
\usepackage[font=small,skip=2pt]{caption}

\def\BibTeX{{\rm B\kern-.05em{\sc i\kern-.025em b}\kern-.08em
    T\kern-.1667em\lower.7ex\hbox{E}\kern-.125emX}}
\begin{document}

\title{\LARGE \bf
H-PAC Hand: Control-Oriented Modeling and Tendon-Elasticity Compensation for an Underactuated Robotic Hand
}

\author{
\IEEEauthorblockN{
Teng Yan$^{1}$, Jiongxu Chen$^{1}$, Teng Wang$^{1}$,
Yue Yu$^{1}$, Qixiang Hua$^{1}$,\\
Zihang Wang$^{2}$, Yongru Chen$^{1}$,
and Bingzhuo Zhong$^{1,*}$
}
\IEEEauthorblockA{
$^{1}$\textit{The Hong Kong University of Science and Technology (Guangzhou)},
Guangzhou, China\\
$^{2}$\textit{Southeast University},
Nanjing, China\\
\{tyan497, jchen074, twang560, yyu704, qhua106, ychan847\}@connect.hkust-gz.edu.cn\\
wzhanggg\_seu@seu.edu.cn;\quad
$^{*}$Corresponding author: bingzhuoz@hkust-gz.edu.cn
}
}

\maketitle

%%%%%%%%%%%%%%%%%%%%%%%%%%%%%%%%%%%%%%%%%%%%%%%%%%%%%%%%%%%%%%%%%%%%%%%%%%%%%%%%
\begin{abstract}
Underactuated tendon-driven hands offer compact actuation and passive compliance, but tendon elongation under restoring-spring loading introduces configuration-dependent joint deviations. This paper presents H-PAC, a modular 6-actuator, 15-DoF robotic hand with a control-oriented modeling and implementation framework. A sparse analytical actuator-joint model is derived from the tendon-routing geometry, and a mechanics-based compensation model is developed to account for tendon-elasticity-induced joint errors.

The proposed method is implemented in a hierarchical architecture: a host computer performs workspace-constrained posture mapping and compensation, while an ESP32 generates synchronized commands for six position-controlled servos. The same control parameters and execution strategy are used across all tasks without task-specific retuning.

Monotonic servo-sweep experiments show that the compensation substantially improves joint-angle prediction. The MAE of the index DIP joint decreases from $1.15^\circ$ to $0.18^\circ$, and all nine evaluated joints achieve an MAE below $0.23^\circ$. Representative postures and grasping configurations are further executed using the same control pipeline without external joint or force sensing in the control loop. The results demonstrate a practical approach to improving posture reproducibility in compact underactuated robotic end-effectors.

\end{abstract}

%%%%%%%%%%%%%%%%%%%%%%%%%%%%%%%%%%%%%%%%%%%%%%%%%%%%%%%%%%%%%%%%%%%%%%%%%%%%%%%%
\section{Introduction}

Compact robotic end-effectors must balance dexterity against actuator count, mass, volume, wiring complexity, and power consumption. Fully actuated dexterous hands provide independent joint control but become increasingly difficult to integrate as the number of DoFs grows. In contrast, underactuated tendon-driven hands coordinate multiple joints with fewer actuators while retaining passive compliance, making them attractive for compact mechatronic systems.

A key limitation of such hands is the repeatability of free-space posture generation. In fingers equipped with passive restoring springs, flexion increases the tendon load, causing elastic elongation that absorbs part of the actuator displacement. This deformation accumulates along the transmission path and produces deviations between geometric predictions and physical joint responses, particularly at distal joints. Although additional joint, displacement, or tendon-tension sensors can compensate for these errors, they increase hardware complexity, wiring, calibration effort, and cost.

\begin{figure}[t]
    \centering
    \includegraphics[width=0.8\columnwidth]{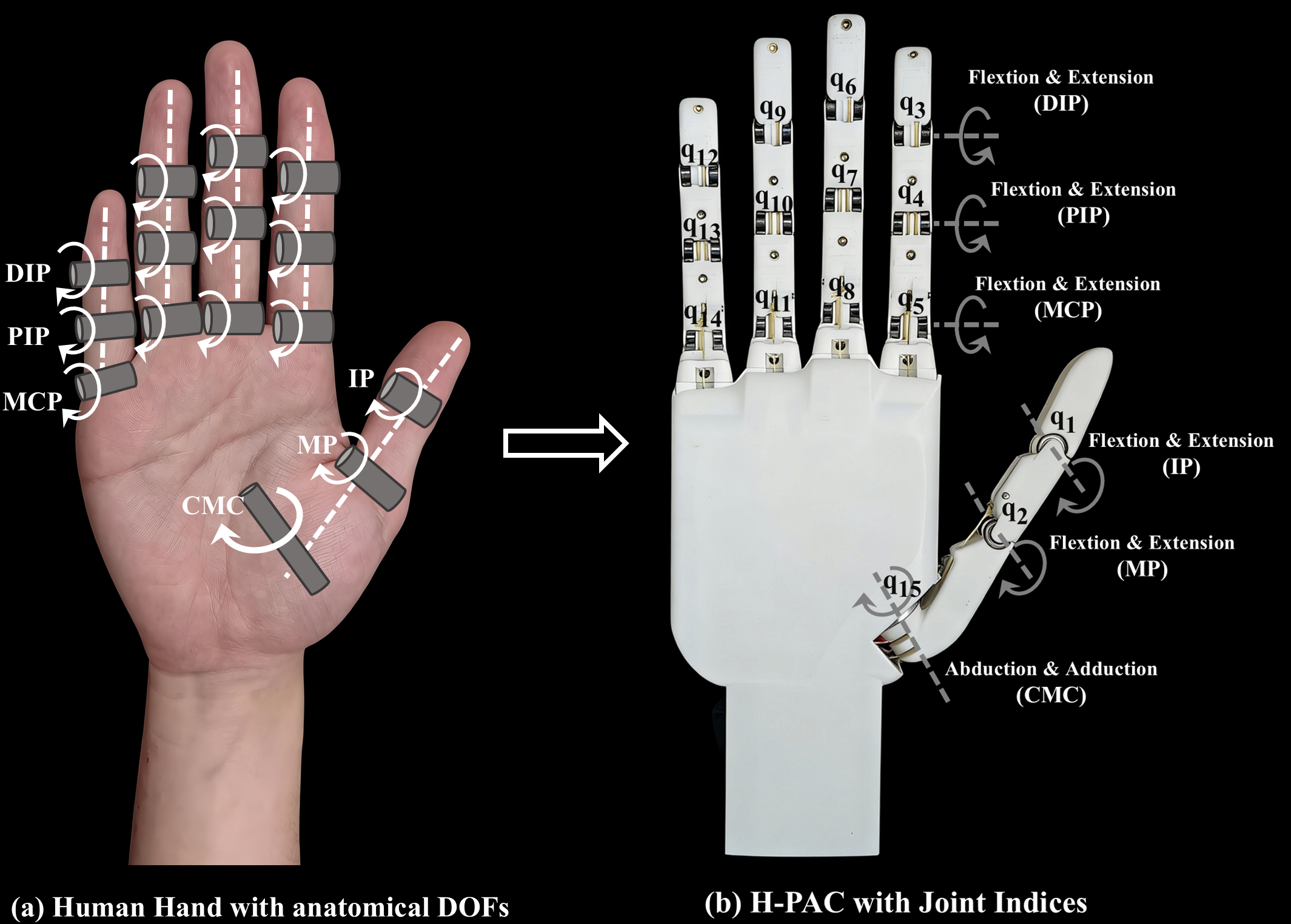}
    \caption{\textbf{Overview and DOF mapping of H-PAC Hand.}  
    (a) Human hand annotated with anatomical DOFs (MCP/PIP/DIP, MP/IP, and CMC).  
    (b) The H-PAC Hand with 15 joint DOFs $q_{1}\!-\!q_{15}$: four long digits each with MCP/PIP/DIP flexion–extension, and the thumb with IP/MP flexion–extension and CMC abduction–adduction.}
    \label{fig:H-PAC_overview}
\end{figure}

To address this problem, this paper presents the \textbf{H-PAC Hand (Hybrid Precision-Augmented Compliance)}, a modular, human-scale tendon-driven robotic hand with 6 actuators and 15 joint DoFs, as shown in Fig.~\ref{fig:H-PAC_overview}. The four long fingers use figure-eight tendon routing to couple their MCP, PIP, and DIP joints. The thumb employs one tendon-driven actuator for coupled IP/MP flexion and one directly coupled actuator for CMC abduction-adduction. Based on this topology, a block-sparse nonlinear actuator-joint model is derived from the tendon-routing geometry, and a mechanics-based compensation model is introduced to account for tendon elongation under restoring-spring loading.

The proposed method is implemented in a hierarchical control architecture. A host computer performs posture generation, workspace constraint handling, analytical mapping, and tendon-elasticity compensation, while an ESP32-based execution layer generates synchronized commands for six position-controlled servos. All target configurations are constrained to the calibrated admissible workspace, and the same communication protocol, motion-duration policy, actuator limits, and command-generation parameters are used across all tasks without task-specific retuning.

The method is evaluated through monotonic servo-sweep experiments by comparing the analytical and compensated models. The compensation reduces the MAE of the index DIP joint from $1.15^\circ$ to $0.18^\circ$, while all nine evaluated joints achieve an MAE below $0.23^\circ$ over the tested ranges. Representative postures and grasping configurations are further executed using the same model, compensation parameters, and embedded control pipeline.

The main contributions are summarized as follows:
\begin{itemize}
    \item \textbf{Sparse actuator-joint modeling:}
    A block-sparse nonlinear model derived from figure-eight tendon geometry provides interpretable posture prediction and command generation on the reachable synergy manifold.

    \item \textbf{Mechanics-based tendon-elasticity compensation:}
    A compensation model describes joint deviations induced by restoring-spring loading and tendon elongation without external joint or force sensing in the control loop.

    \item \textbf{Hierarchical embedded implementation and validation:}
    A host-computer and ESP32 control architecture integrates workspace-constrained command generation and synchronized servo execution, validated through servo-sweep experiments and representative posture and grasping demonstrations.
\end{itemize}

\section{Related Work}

\subsection{Motivation for Compact and Underactuated Designs}

Fully actuated dexterous hands assign an independent actuator to each DoF, providing high expressivity and control bandwidth, as demonstrated by the Gifu Hand series \cite{Mouri2011,Kawasaki2002} and the DLR Hand Arm System \cite{Grebenstein2012}. However, their weight, volume, power consumption, and cost increase substantially with the number of DoFs \cite{Grebenstein2012,Tokunaga2025}. Compact and underactuated designs therefore coordinate multiple joints using fewer actuators through mechanical coupling, tendon routing, or elastic elements \cite{Piazza2019,Brown2007,Kim2021,Schunk2025,Bridgwater2012,Wall2017}. Although these systems improve deployability, safety, grasp robustness, and compliant interaction \cite{He2019}, their reduced actuation introduces challenges in joint-space modeling and calibration.

\subsection{Mapping and Calibration Methods}

Early studies commonly used empirical linear scaling or synergistic models to map actuator strokes to finger joint angles \cite{Telegenov2015}. Low-dimensional synergy projections of human hand motion reduce the control dimension and enable efficient but approximate posture representation \cite{Dwivedi2020,Romero2013,Santos2025,Liu2008}. However, empirical mappings are sensitive to manufacturing tolerances and assembly variations, limiting their transferability across prototypes.

Analytical approaches instead derive closed-form tendon-length joint-angle relationships from transmission geometry and equivalent-radius formulations. These models provide interpretable and reusable actuator-joint mappings \cite{kim2020joint,sainul2016three}. Nevertheless, practical effects such as tendon guiding, wrapping, and slack can produce systematic deviations, particularly over large joint ranges, unless appropriate calibration or compensation is introduced.

\subsection{Sensing and Data-Driven Compensation}

High-fidelity calibration often relies on tension, displacement, or optical sensors to identify elastic deformation, friction, and hysteresis in tendon transmissions \cite{Liu2008,Johansson2009,Dargahi2004}. Although physically interpretable, such methods increase hardware and calibration complexity. End-to-end learning has also been used to model actuator-to-joint or actuator-to-end-effector relationships \cite{Pinto2016,Kalashnikov2018}. These approaches offer strong task adaptability but typically require substantial data and computation, while providing limited interpretability and cross-prototype transferability \cite{Mohammed2020}.

Augmenting an analytical mapping with low-order, small-sample, angle-dependent compensation therefore provides a practical balance between physical transparency and engineering usability. H-PAC follows this direction by integrating analytical geometric
mapping with mechanics-based tendon-elasticity compensation on a
6-actuator, 15-DoF platform. The resulting framework enables sub-degree posture reproducibility without additional sensing while retaining the task versatility of an underactuated dexterous hand.

\section{methodology}
\label{metho}
\subsection{Hand design and modeling}

\begin{figure}[t]
    \centering
    \includegraphics[width=1.0\columnwidth]{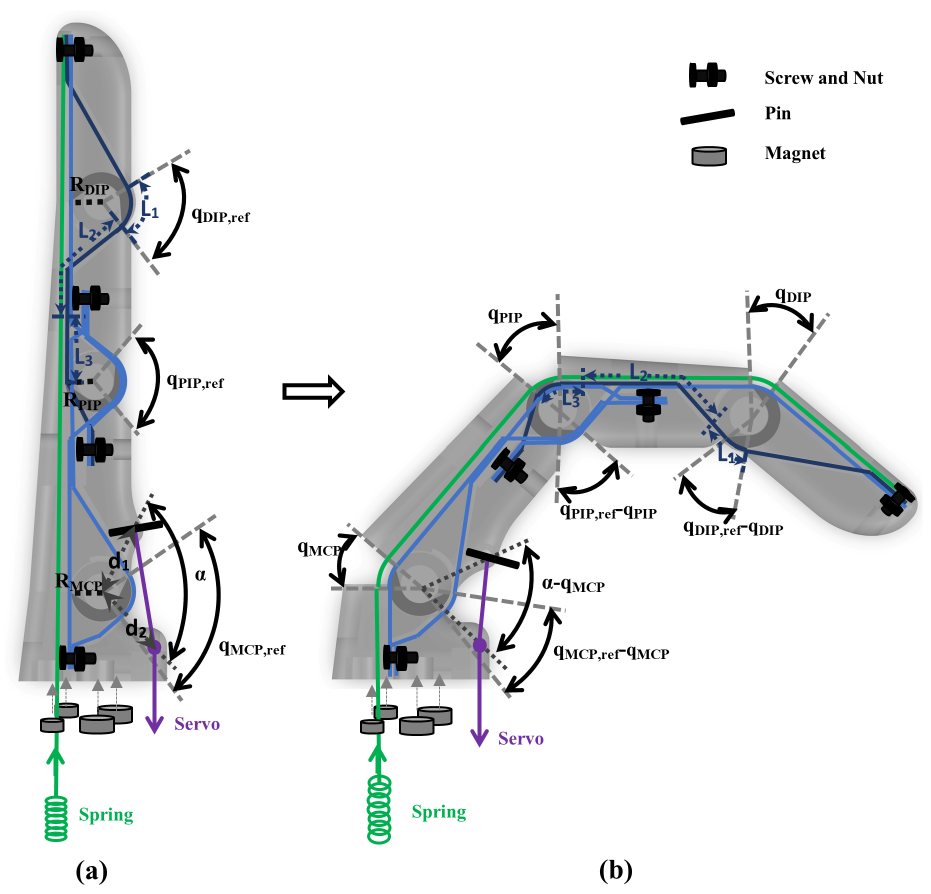}
    \caption{\textbf{Tendon routing and analytic modeling for four fingers.}
    Four-finger figure-eight routing from full extension to an intermediate flexion: straight segments $L_k$, joint pulleys of radius $R_i$, cables connected to the servo hubs (purple), the passive "cable-spring" return path (green), parameters related to the closed-form geometric expression of the MCP joint ($\alpha$, $d_{1}$, $d_{2}$), and joint angles $q_{\text{MCP}}$, $q_{\text{PIP}}$, $q_{\text{DIP}}$.  $q_{i,\text{ref}}$ (where $i=\{\text{DIP},\text{PIP},\text{MCP}\}$) denotes the structural geometric limits of the independent finger module for analytical derivation, while the actual operational range limits are constrained by the palm housing to prevent excessive flexion.}
    \label{fig:fingers}
\end{figure}

The H-PAC Hand adopts a 6-actuator, 15-DoF tendon-driven topology. Servos 2-5 independently actuate the index, middle, ring, and pinky fingers, each using figure-eight tendon routing across the DIP, PIP, and MCP joints to produce coupled flexion, as shown in Fig.~\ref{fig:fingers}. Servo 1 drives the coupled IP/MP flexion of the thumb, while servo 6 directly actuates its CMC abduction-adduction joint, as shown in Fig.~\ref{fig:thumb}.

Additional hardware features include:
\begin{itemize}
\item \textbf{Passive Cable-Spring Return Mechanism:}
A passive cable-spring path restores each finger to full extension when tendon tension is released.

\item \textbf{Low-Friction Biomimetic Design:}
The fingers follow 1:1 human-hand proportions and are arranged with adjacent $5^\circ$ offsets to reproduce the natural palmar arc. Stepped metacarpal heads and joint bearings reduce friction.

\item \textbf{Human-Like Joint Ratio Design:}
The DIP:PIP:MCP pulley-radius ratio is $5:4:5$, yielding an ideal joint-angle ratio of $4:5:4$. The thumb IP:MP radius ratio is $5:4$, producing a $4:5$ angle ratio. Details are given in Sections~\ref{Four-Finger Model} and~\ref{Thumb Model}.
\end{itemize}

Following notations are adopted for subsequent analysis:

\begin{itemize}
    \item \textbf{$\mathbf{u}=[u_{1}, \dots, u_{6}]^\top \in \mathbb{R}^{6}$}: servo angles (rad); 
    $\mathbf{r} = [r_{1}, \dots, r_{6}]^\top$ are the corresponding pulley radii (m).  

    \item $\mathbf{q} = [q_{1}, \dots, q_{15}]^\top \in \mathbb{R}^{15}$: joint angles (rad), indexed as in Fig.~\ref{fig:H-PAC_overview};
    $\mathbf{R} = [R_1, \dots, R_{14}]^\top$: joint radii (m); $R_{15}$
 is not used because the CMC joint is directly driven.

    \item \textbf{$\mathbf{H} \in \mathbb{R}^{15 \times 6}$}: primary mapping matrix (block-sparse, with only 15 non-zero entries; each non-zero entry is a function of a single $u_j$). 

    \item \textbf{$\mathbf{b(q)} = [b_1(q_1), \dots, b_{15}(q_{15})]^\top \in \mathbb{R}^{15}$}: compensation term (rad), defined as the difference between the ideal joint angle $\mathbf{q_{\text{ideal}}}$ and the compensated joint angle $\mathbf{q_{\text{comp}}}$, i.e., $\mathbf{b(q)} = \mathbf{q_{\text{ideal}}} - \mathbf{q_{\text{comp}}}$; the compensation term is only non-zero for interphalangeal joints (IP, PIP, DIP) excluding MCP and thumb MP joints.

    \item \textbf{$k_s$ (N/m)}: stiffness of the restoring spring; \textbf{$EA$ (N)}: axial stiffness of the tendon, where $E$ is the Young's modulus of the tendon material and $A$ is the cross-sectional area.
\end{itemize}

\subsection{Single-Finger Model}
\subsubsection{Four-Finger Model}\

\label{Four-Finger Model}
\textbf{Tendon length conservation under the figure-eight routing.} 
Due to the figure-eight routing of tendons between adjacent joints, the extension and contraction of the tendon segments at neighboring joints mutually compensate during flexion. According to the geometric constraints, the incremental tendon lengths at each joint satisfy $R_{\mathrm{DIP}} \Delta q_{\mathrm{DIP}} = R_{\mathrm{PIP}} \Delta q_{\mathrm{PIP}}$ and $R_{\mathrm{PIP}} \Delta q_{\mathrm{PIP}} = R_{\mathrm{MCP}} \Delta q_{\mathrm{MCP}}$. Consequently, in an ideal state neglecting initial tension, the joint angles follow the coupled relationship
\[
R_{\mathrm{DIP}}\,q_{\mathrm{DIP}}
=R_{\mathrm{PIP}}\,q_{\mathrm{PIP}}
=R_{\mathrm{MCP}}\,q_{\mathrm{MCP}}.
   \tag{1}\]
By setting $R_{\mathrm{DIP}} : R_{\mathrm{PIP}} : R_{\mathrm{MCP}} = 5:4:5$ , the joints exhibit an equal-angle synergy: 
\[
{\;q_{\mathrm{DIP}}:q_{\mathrm{PIP}}:q_{\mathrm{MCP}}=4:5:4 .\;}  \tag{2}
\]

\textbf{Closed-form expression of the MCP joint.}
As shown in Fig.~\ref{fig:fingers}(a), (b), let $d_1,d_2$ denote the fixed distances from the guides to the MCP rotation center, and let $\alpha$ be the initial included angle. The initial chord length is  
\[
L_0=\sqrt{d_1^2+d_2^2-2d_1d_2\cos\alpha}. \tag{3}
\]
The tendon retraction by servo $j$ is $\Delta L=r_j u_j$, giving current chord length $L=L_0-\Delta L$.  
By the cosine law, the instantaneous angle is given by
\[
\varphi=\arccos\!\left(\frac{d_1^2+d_2^2-L^2}{2d_1 d_2}\right). \tag{4}
\]
Flexion is defined as a decrease of the included angle from its initial value $\alpha$, the MCP rotation is  
\[
{\;q_{\mathrm{MCP}}=\alpha-\arccos\!\left(\frac{d_1^2+d_2^2-(L_0-r_j u_j)^2}{2d_1d_2}\right).\;}
 \tag{5}\]

\begin{figure}[t]
    \centering
    \includegraphics[width=1.0\columnwidth]{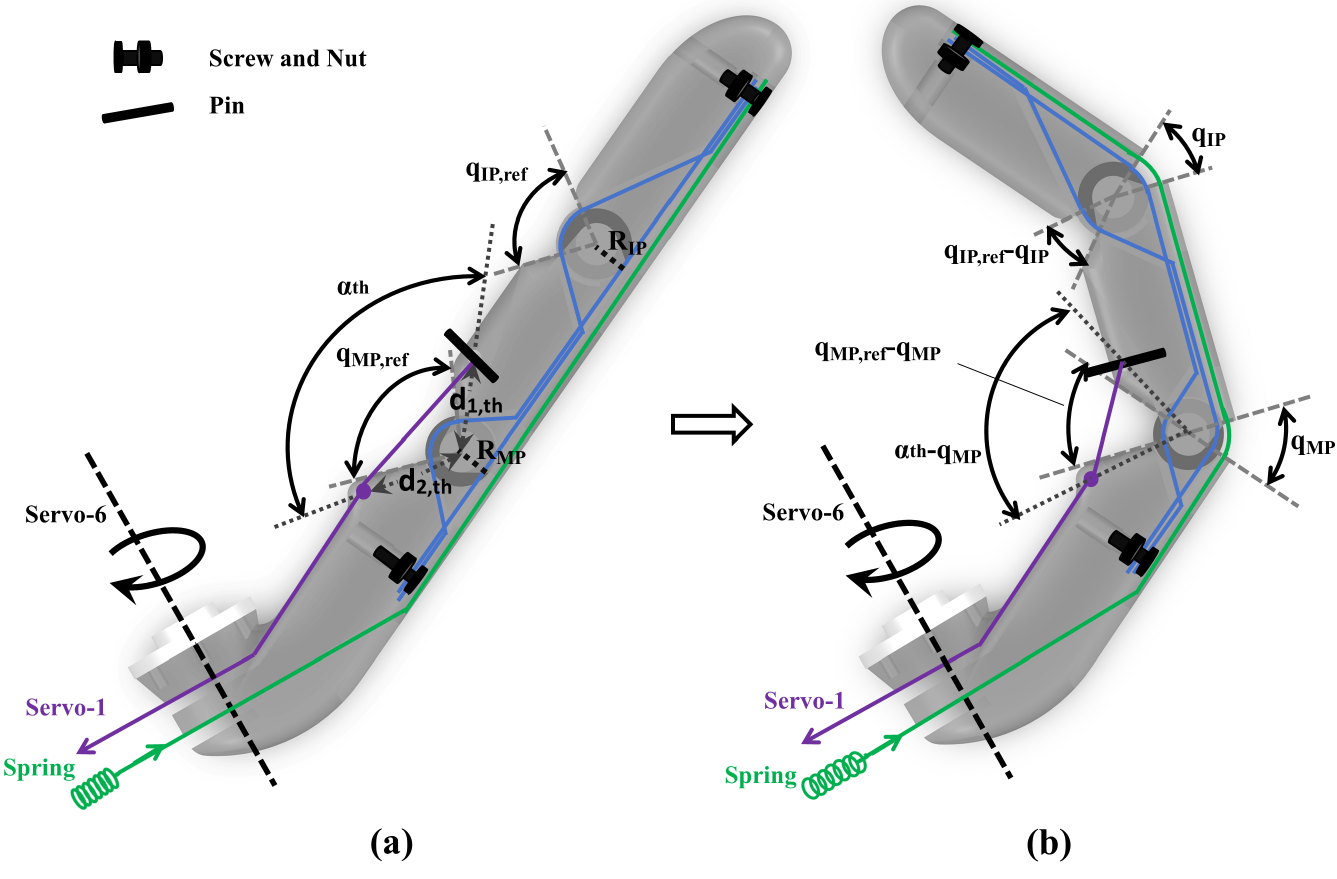}
    \caption{\textbf{Tendon routing and analytic modeling for thumb.}
    The thumb uses figure-eight tendon routing for coupled MP/IP flexion, parameterized by $\alpha_{\mathrm{th}}$, $d_{1,\mathrm{th}}$, and $d_{2,\mathrm{th}}$, with joint angles $q_{\mathrm{MP}}$ and $q_{\mathrm{IP}}$. The structural limits $q_{i,\mathrm{ref}}$ ($i=\{\mathrm{IP},\mathrm{MP}\}$) are defined analogously to the four-finger model. Servo-6 drives the CMC joint, with its output axis coaxial to the CMC joint axis and the servo horn directly rigidly attached to the CMC joint.}

    \label{fig:thumb}
\end{figure}

\subsubsection{Thumb Model}\

\label{Thumb Model}
\textbf{Coupled IP–MP Flexion.} 
Fig.~\ref{fig:thumb} illustrates the thumb moving from extension to an intermediate flexion. Two tendons at IP/MP joints are also routed in a figure-eight configuration, with a design ensuring $R_{\mathrm{IP}}:R_{\mathrm{MP}}=5:4$, so  
\[
{\;q_{\mathrm{IP}}:q_{\mathrm{MP}}=4:5 .\;}
 \tag{6}\]
Let the thumb guiding parameters be $d_{1,\mathrm{th}}, d_{2,\mathrm{th}}, \alpha_{\mathrm{th}}$, as in Fig.~\ref{fig:thumb}(a). Analogous to the finger MCP case, applying the cosine law yields,  
\begin{align*}
q_{\mathrm{MP}} &= \alpha_{\mathrm{th}}
 - \arccos\!\left(
   \frac{d_{1,\mathrm{th}}^2+d_{2,\mathrm{th}}^2-(L_{0,\mathrm{th}}-r_1u_1)^2}
        {2d_{1,\mathrm{th}}d_{2,\mathrm{th}}}
   \right). \tag{7}
\end{align*}

\textbf{CMC abduction.}
The abduction of the CMC joint is implemented via a direct drive mechanism. The proximal end of the thumb's PLP (Proximal Phalanx) is coupled directly to the servo output axis. The relationship between the joint angle $q_{CMC}$ and the servo input $u_6$ is expressed as
\[
{\;q_{\mathrm{CMC}}=g_{\mathrm{CMC}}\,u_6,\quad g_{\mathrm{CMC}}=1.\;}  \tag{8}
\]

\subsection{Sparse Actuator-to-Joint Mapping}
By aggregating the relations of four fingers and the thumb, a \(15 \times 6\) function-type mapping matrix $\mathbf{H}(\mathbf{u})$ is obtained:

\begin{itemize}
    \item Four-finger block (each $3 \times 1$): the three rows corresponding to DIP/PIP/MCP of the same finger share the same functional term 
  $\mathbf{H}_{ij}(u_j) = \tfrac{q_{\mathrm{finger}}}{u_j}$; 

  \item Thumb flexion block ($2 \times 1$): $\mathbf{H}_{i1}(u_1) = \tfrac{q_{\mathrm{thumb}}}{u_1}$;

  \item CMC block ($1 \times 1$): $\mathbf{H}_{15,6} = g_{\mathrm{CMC}} = 1$ (constant).  

\end{itemize}

Thus, the preliminary mapping relation is expressed as
\[
\mathbf{q_{ideal}} = \mathbf{H}(\mathbf{u})\,\mathbf{u}. \tag{9}
\]

The forward mapping predicts the reachable joint posture from the six actuator commands. Since the hand is underactuated, an arbitrary 15-dimensional posture cannot be independently commanded. Target postures are therefore restricted to the reachable synergy manifold, and actuator commands are obtained using the blockwise inverse relations or a pseudoinverse-based least-squares solution.

\subsection{Mechanics-Based Tendon-Elasticity Compensation}
\subsubsection{Four-Finger Model}\

While the joint angles theoretically follow a $4:5:4$ coupling ratio, the introduction of a fingertip restoring spring for self-extension significantly impacts the transmission accuracy in practical applications. As the finger curls, the spring tension $T_{spr}$ acts upon the entire transmission chain, inducing non-negligible elastic strain in the driving and coupling cables. This strain accumulates along the transmission path, causing actual joint angles to deviate from ideal values, with more pronounced discrepancies observed in distal joints.

According to Hooke's Law, the restoring force generated by the spring is proportional to the total excursion:
\[T_{spr} = k_s (R_{\mathrm{DIP}}\,q_{\mathrm{DIP}}
+ R_{\mathrm{PIP}}\,q_{\mathrm{PIP}}
+ R_{\mathrm{MCP}}\,q_{\mathrm{MCP}}). \tag{10}\]

The elastic elongation $\delta$ of the cable for each segment follows the fundamental equation $\delta = \frac{T L}{EA}$. Due to this deformation, the displacement intended for the driven joints is partially absorbed. Since the MCP joint is located closest to the actuator and has the shortest effective transmission path, its elastic displacement is assumed to be small relative to that of the downstream PIP and DIP joints, leading to $q_{comp,\mathrm{MCP}} \approx q_{ideal,\mathrm{MCP}}$. For the PIP and DIP joints, the angular deviation $b(q_j)$ between their compensated actual angles $q_{comp,j}$ and ideal angles $q_{ideal,j}$ is defined as
 \[b(q_j) = \frac{\delta_j}{R_j} = \frac{T_{spr} \cdot \sum L_{path}}{EA \cdot R_j},  \tag{11} \]
where $ j=\{\text{PIP},\text{DIP}\} $, and the actual joint angle is given by $q_{comp,j}=q_{ideal,j} - b(q_j)$.

Integrating the aforementioned constraints, the system is modeled as a set of equilibrium equations. By incorporating the elastic deviation terms, the matrix equation is established:
\[\mathbf{M} \mathbf{q_{comp}} = \mathbf{A} q_{comp,\mathrm{MCP}}.   \tag{12}\]

where $\mathbf{A} = [1, 1.25, 1]^T$ is the ideal coupling vector determined by the hardware geometry. The system coupling matrix $\mathbf{M}$ is defined as
\vspace{-2pt}
$$\mathbf{M} = \begin{bmatrix} 1 & 0 & 0 \\ C_\mathrm{PIP} R_\mathrm{MCP} & 1 + C_\mathrm{PIP} R_\mathrm{PIP} & C_\mathrm{PIP} R_\mathrm{DIP} \\ C_\mathrm{DIP} R_\mathrm{MCP} & C_\mathrm{DIP} R_\mathrm{PIP} & 1 + C_\mathrm{DIP} R_\mathrm{DIP} \end{bmatrix},$$
where $C_\mathrm{PIP} = \frac{k_s L_{\mathrm{MCP},\mathrm{PIP}}}{EA R_\mathrm{PIP}}$ and $C_\mathrm{DIP} = \frac{k_s (L_{\mathrm{MCP},\mathrm{PIP}} + L_{\mathrm{PIP},\mathrm{DIP}})}{EA R_\mathrm{DIP}}$ represent the cumulative structural compliance (Let $L_{\mathrm{MCP},\mathrm{PIP}}$ and $L_{\mathrm{PIP},\mathrm{DIP}}$ be the path lengths between MCP-PIP and PIP-DIP joints, respectively). By inverting $\mathbf{M}$, the compensated joint-response prediction is obtained as $\mathbf{q_{comp}} = \mathbf{M}^{-1} \mathbf{A} q_{comp,\mathrm{MCP}}$, enabling physical compensation for flexible transmission errors.

\subsubsection{Thumb Model}\

Since the flexion and extension of the thumb involve only two active joints ($q_\mathrm{MP}$ and $q_\mathrm{IP}$), its mechanics-based compensation model can be formulated as a second-order reduced version of the generalized model described above.

the system coupling matrix $\mathbf{M}$ is simplified into a $2 \times 2$ form. The matrix equation is expressed as

\[\begin{bmatrix} 1 & 0 \\ C R_\mathrm{MP} & 1 + C R_\mathrm{IP} \end{bmatrix} \begin{bmatrix} q_{comp,\mathrm{MP}} \\ q_{comp,\mathrm{IP}} \end{bmatrix} = \begin{bmatrix} 1 \\ 0.8 \end{bmatrix} q_{comp,\mathrm{MP}},   \tag{13}\]
where $C = \frac{k_{s} L_{\mathrm{MP},\mathrm{IP}}}{EA R_\mathrm{IP}}$ is the compliance compensation coefficient for the thumb.

Since the CMC joint of the thumb is directly coupled to and controlled by the servo-6 output axis, the transmission chain is remarkably short and free of significant elastic load. Therefore, no mechanical compensation is required.

\begin{table}[b]
\centering
\caption{\textbf{Admissible workspace of each finger joint of H-PAC.} A monotonic servo sweep dedicated to workspace
determination is used to obtain $[\mathbf{q}_{\min},
\mathbf{q}_{\max}]$.}
\label{tab:joint_workspace}
\begin{tabular}{lcc}
\hline
Finger Joint & Joint Label & \makecell[c]{Admissible Workspace \\ $[q_{\min}, q_{\max}]$ (rad)} \\
\hline
Thumb IP     & $q_1$       & $[0.00, 0.99]$ \\
Thumb MP     & $q_2$       & $[0.00, 1.25]$ \\
Index DIP    & $q_3$       & $[0.00, 1.31]$ \\
Index PIP    & $q_4$       & $[0.00, 1.61]$ \\
Index MCP    & $q_5$       & $[0.00, 1.27]$ \\
Middle DIP   & $q_6$       & $[0.00, 1.28]$ \\
Middle PIP   & $q_7$       & $[0.00, 1.58]$ \\
Middle MCP   & $q_8$       & $[0.00, 1.24]$ \\
Ring DIP     & $q_9$       & $[0.00, 1.29]$ \\
Ring PIP     & $q_{10}$    & $[0.00, 1.59]$ \\
Ring MCP     & $q_{11}$    & $[0.00, 1.25]$ \\
Pinky DIP    & $q_{12}$    & $[0.00, 1.25]$ \\
Pinky PIP    & $q_{13}$    & $[0.00, 1.55]$ \\
Pinky MCP    & $q_{14}$    & $[0.00, 1.23]$ \\
Thumb CMC    & $q_{15}$       & $[0.00, 1.57]$ \\
\hline
\end{tabular}
\end{table}

All parameters used in the analytical and compensation models are obtained from the physical design or material properties. Because of geometric differences among the assembled fingers, each long finger uses its own parameter set rather than sharing a common set of parameters.

\begin{figure*}[t]
    \centering
    \includegraphics[width=0.8\textwidth]{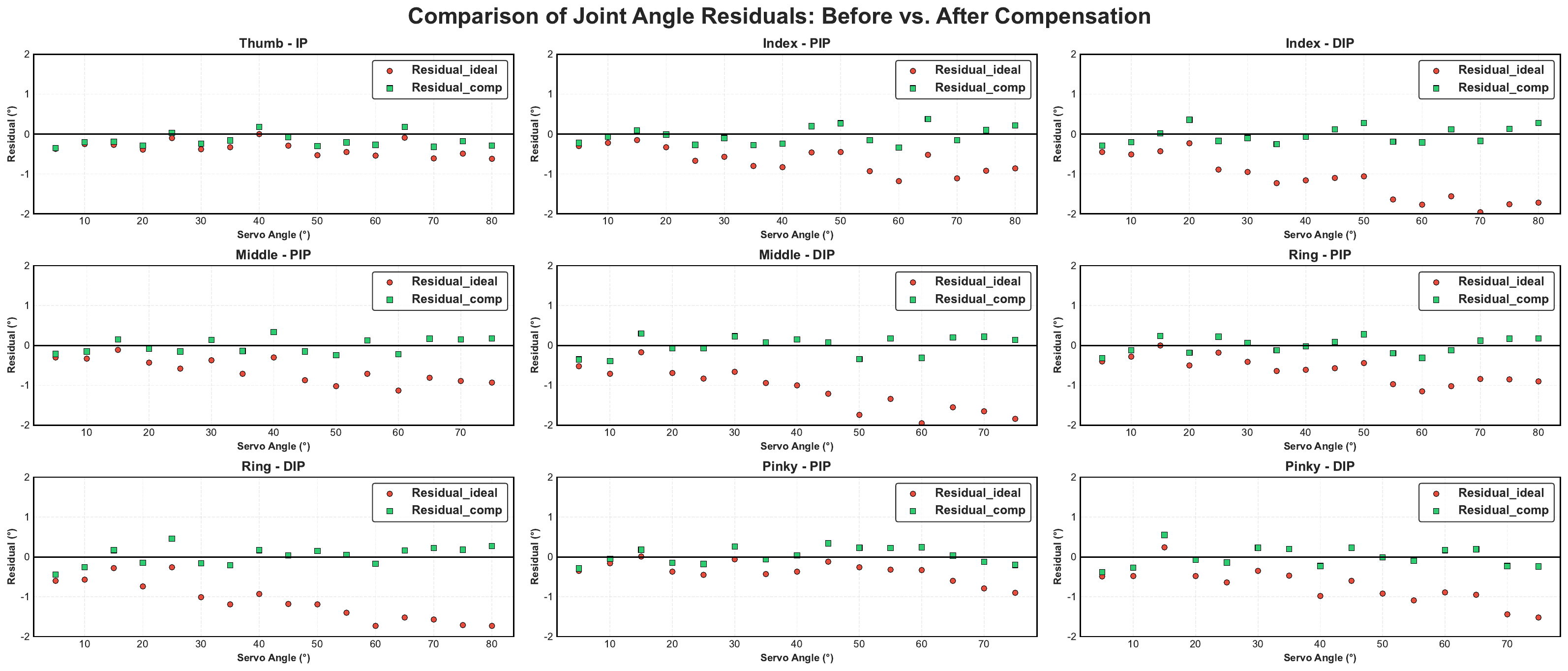}
    \caption{\textbf{Comparative analysis of joint angle residuals under uncompensated and compensated control.}
    This figure illustrates the relationship between the angle residuals and the servo control angles for target joints across the five fingers. Red circles represent the original uncompensated residuals (Residual\_{ideal}) and green squares denote the residuals after mechanics-based compensation (Residual\_{comp}). All subplots are standardized with a Y-axis range from $-2.0$ to $+2.0$ degrees to facilitate a direct comparison of performance.}
    \label{fig:finger_residual_plot}
\end{figure*}

\begin{figure*}[t]
    \centering
    \includegraphics[width=0.7\textwidth]{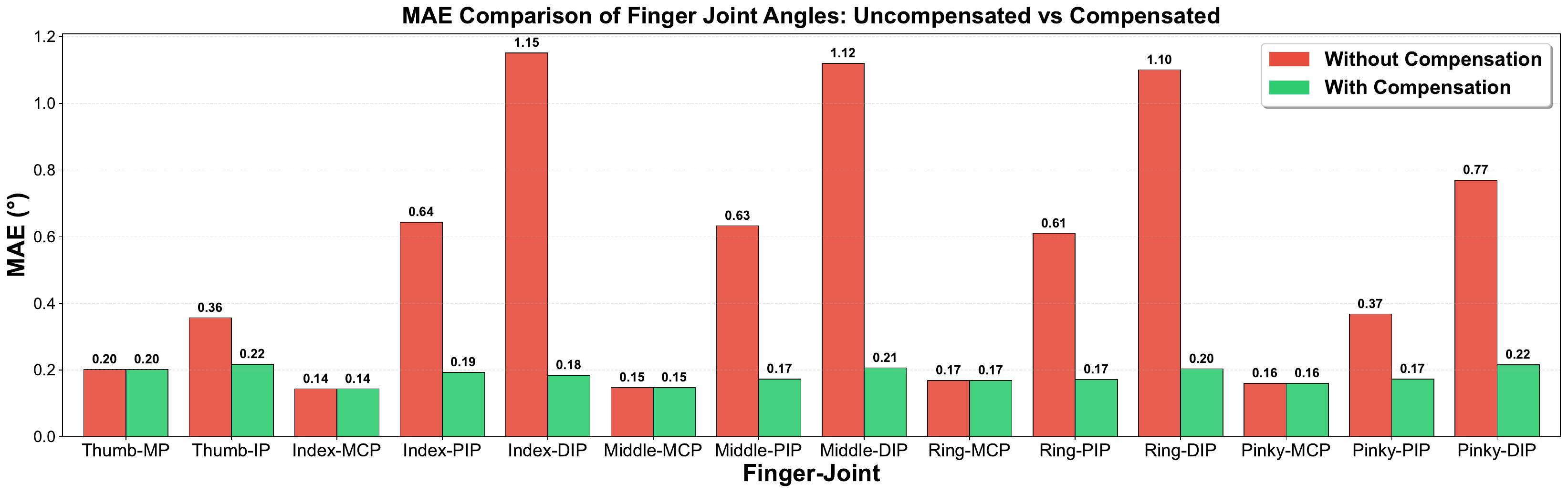}
    \caption{\textbf{Mean Absolute Error (MAE) comparison of joint angles for five fingers of H-PAC Hand under uncompensated and compensated conditions.} The red bars represent the MAE between real joint angles and ideal joint angles (without compensation), while the green bars denote the MAE between real joint angles and compensation joint angles (with compensation). The fingers include the thumb (with MP and IP joints) and four fingers (each with MCP, PIP and DIP joints). The MAE is calculated in the unit of degrees ($^\circ$).}
    \label{fig:finger_mae_plot}
\end{figure*}

\subsection{Hierarchical Control and Safe Motion Execution}

The H-PAC Hand adopts a hierarchical architecture comprising a host computer, an ESP32-based execution layer, and six servo actuators. The host computer performs target-posture generation, joint-space constraint handling, actuator-joint mapping, and tendon-elasticity compensation. The ESP32 parses control commands and executes synchronized low-level motion, while each servo closes its internal position-control loop.

Given a desired posture $\mathbf{q}_d$, the target joint angles are first constrained to the calibrated admissible workspace in Table~\ref{tab:joint_workspace}:
\[
\tilde{\mathbf{q}}_d =
\operatorname{clip}
\left(
\mathbf{q}_d,
\mathbf{q}_{\min},
\mathbf{q}_{\max}
\right),
\tag{14}
\]
where clipping is applied jointwise to keep the command within the calibrated operating range and reduce the risk of excessive tendon elongation, joint interference, and mechanical jamming. The compensated actuator command is then computed as
\[
\mathbf{u}^{\star}
=
\mathbf{H}^{+}
\left[
\tilde{\mathbf{q}}_d+
\mathbf{b}
\left(
\tilde{\mathbf{q}}_d
\right)
\right],
\tag{15}
\]
where $\mathbf{H}^{+}$ is the command-generation operator associated with the sparse actuator-joint mapping, and $\mathbf{b}(\cdot)$ is the tendon-elasticity compensation term.

The host computer transmits a control packet containing six target servo angles, a prescribed motion duration, and a checksum to the ESP32. After validation and parsing, the ESP32 generates the corresponding PWM signals. All actuators use the same motion duration, enabling synchronized finger motion while reducing abrupt tendon-load changes and mechanical impact during posture transitions.

The servos operate in position-control mode with internally closed position loops. No additional joint-angle or force-feedback control is implemented by the ESP32; therefore, the analytical mapping and elasticity compensation form a model-based feedforward layer. All experiments use the same communication protocol, PWM scheme, motion-duration policy, actuator limits, and command-generation parameters without task-specific retuning. Accordingly, all postures and grasps in Fig.~\ref{fig:table} are executed using the same control pipeline.

\section{Experiments and Analysis}
\subsection{Experimental Setup}

The experimental platform is the H-PAC Hand described in \ref{metho}.
All angles are represented in radians unless otherwise specified. External measurements are used exclusively for offline evaluation (i.e., not involved in closed-loop control). Each phalanx is augmented with a visible rigid reference (fiducial and edge model). Using a calibrated vision pipeline, we reconstruct the angle of each joint (single-plane flexion–extension), denoted as $\mathbf{q_{\text{meas}}}$. To ensure consistency with the control layer, all recorded values are first converted to radians.
The sampled sequences are then processed with a second-order zero-phase Butterworth low-pass filter (cutoff frequency set according to motion bandwidth), followed by unwrapping and drift correction.

Two monotonic servo sweeps are conducted for different purposes. The first sweep is used exclusively to determine the admissible joint workspace $[\mathbf{q}_{\min},\mathbf{q}_{\max}]$ reported in Table~\ref{tab:joint_workspace}. A separate sweep is then performed for model evaluation, during which $\mathbf{q}_{\mathrm{meas}}$ is recorded and compared with the analytical and compensated predictions. The measurements from this evaluation sweep are not used to identify or fit the physical parameters of the compensation model.

\begin{figure*}[t]
    \centering
    \includegraphics[width=0.7\textwidth]{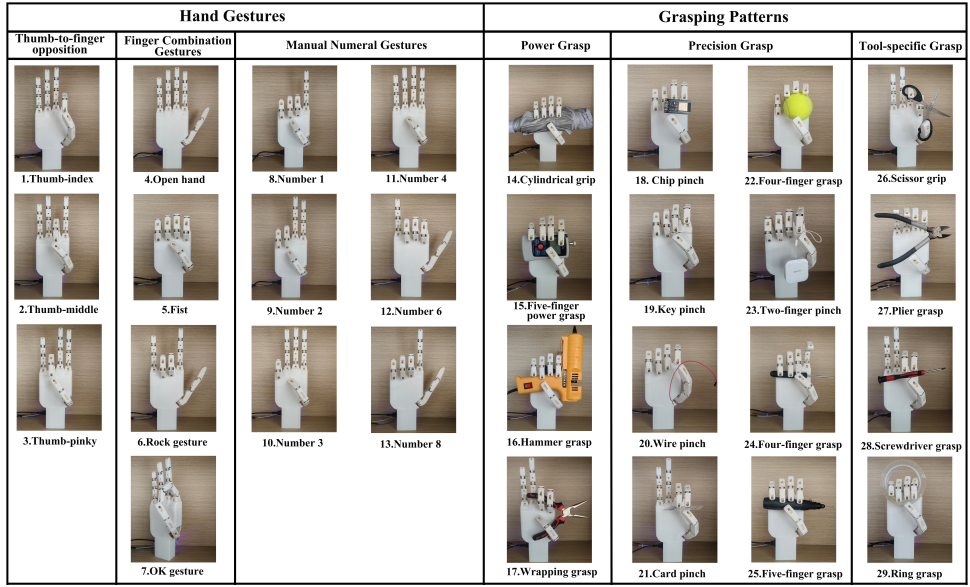}
    \caption{\textbf{Representative posture and grasping demonstrations of the H-PAC Hand.}
    Left: Hand gestures, including precise thumb-to-finger opposition (thumb opposing the index, middle, and ring fingers respectively), representative multi-finger combinations and digit gestures 1-8. Right: An extended set of 16 representative grasps under the same controller, covering diverse contact topologies (enveloping, lateral, and tip pinches) and tool morphologies (rod-like, ring-like, and press-type).
    All results from the same control stack (analytic mapping + mechanics-based compensation), without external joint or force sensing in the control loop.}
    \label{fig:table}
\end{figure*}

\subsection{Mapping Accuracy via Servo Sweeps}

\textbf{Objective.}
This experiment evaluates the accuracy and residual patterns of the analytical mapping with mechanics-based compensation against the analytical-only baseline.

\textbf{Method.}
For each servo $j$, a separate  monotonic sweep over its admissible range is discretized into $K_j$ sampling points. At each point, the joint angles $\mathbf{q}_{\mathrm{meas}}$ are recorded. The analytical and compensated predictions are respectively computed as
\[
\mathbf{q}_{\mathrm{ideal}}
=
\mathbf{H}(\mathbf{u})\mathbf{u},
\qquad
\mathbf{q}_{\mathrm{comp}}
=
\mathbf{H}(\mathbf{u})\mathbf{u}
-
\mathbf{b}(\mathbf{q}),
\]
and compared with $\mathbf{q}_{\mathrm{meas}}$.

\textbf{Evaluation Metrics.}
The angular residual is defined as
$\delta q=q_{\mathrm{meas}}-q_{\mathrm{model}}$ and describes the error variation over the actuation range. Fig.~\ref{fig:finger_residual_plot} presents the residuals of nine target joints, including the thumb IP and the PIP/DIP joints of the four long fingers. The mean absolute error (MAE) summarizes the prediction accuracy of the analytical and compensated models, as shown in Fig.~\ref{fig:finger_mae_plot}.

%\textbf{Results and Analysis.} 
%(i) As shown in Fig.~\ref{fig:finger_residual_plot}, the pre-compensation residuals exhibit distinct non-linear fluctuations and systematic offsets, with deviations approaching 2.0$^\circ$ in certain joints. After compensation, the residuals fluctuate closely around the $0^\circ$ baseline and maintain a uniform distribution across the entire motion range without significant trending bias. (ii) The Mechanics-Based Compensation model achieves a significant reduction in the MAE for the target joints (IP, PIP, DIP, excluding MCP and thumb MP). Among all target joints, the index finger DIP joint exhibits the most remarkable compensation effect, with its MAE decreasing sharply from 1.15 to 0.18. Meanwhile, the thumb IP joint and the PIP/DIP joints of other fingers also show obvious MAE reductions, which verifies the effectiveness of the proposed compensation model. (iii) After compensation, the MAE of all target joints is $\leq 0.23$, which falls within the sub-degree level. This indicates that the compensated joint angle predictions meet the high-precision requirements for practical application.

\textbf{Results and Analysis.}
As shown in Fig.~\ref{fig:finger_residual_plot} and Fig.~\ref{fig:finger_mae_plot}, the uncompensated residuals exhibit nonlinear variations and systematic offsets, approaching $2.0^\circ$ for some joints. After compensation, the residuals remain closer to the $0^\circ$ baseline with reduced variation over the motion range. The compensation lowers the MAE of the evaluated IP, PIP, and DIP joints, with the index DIP showing the largest reduction from $1.15^\circ$ to $0.18^\circ$. The thumb IP and the remaining PIP/DIP joints also exhibit clear improvements. After compensation, the MAE of all nine target joints remains below $0.23^\circ$, demonstrating sub-degree joint-angle prediction accuracy.

\subsection{Static Posture Expressibility}

Static posture expressibility is evaluated using 13 target configurations shown in Fig.~\ref{fig:table}(1-13). These include: (i) thumb opposition to the index, middle, and ring fingers (Fig.~\ref{fig:table}(1-3)); (ii) representative multi-finger gestures (Fig.~\ref{fig:table}(4-7)); and (iii) digit gestures 1-8 (Fig.~\ref{fig:table}(8-13)). Each target posture $\mathbf{q}_d$ is derived from reference human-hand geometry and mapped to the joint space according to the indexing in Fig.~\ref{fig:H-PAC_overview}.

%\subsection{Static Posture Expressibility}
%This experiment evaluates the expressibility of static postures using 13 target configurations, as illustrated in Fig.~\ref{fig:table}(1-13). The set includes three categories: (i) opposition of the thumb to the index, middle, ring fingers (Fig.~\ref{fig:table}(1–3)); (ii) representative gestures made by multi-finger coordination (Fig.~\ref{fig:table}(4–7)); and (iii) digit gestures 1–8 (Fig.~\ref{fig:table}(8–13)).  

%Each target posture $\mathbf{q_d}$ is derived from reference human hand geometry and mapped to the joint space following the indexing defined in Fig.\ref{fig:H-PAC_overview}.

\subsection{Grasp Diversity}

Grasp diversity is evaluated using representative power, precision, and tool-specific grasps. The target posture $\mathbf{q}_d$, derived from reference human-hand shapes, is executed through the control command $\mathbf{u}^{\star}$. For tasks involving reach, closure, lift, and place transitions, all phases use the same model-based command-generation method and the servos' internal position-control mode.
The evaluation focuses on: (i) thumb-finger opposition and lateral pinch enabled by the single-actuated fingers and dual-actuated thumb; (ii) grasp stability across objects of different sizes and weights; and (iii) tool-oriented posture shaping, particularly the role of the CMC joint in lateral motion and opposition.

As shown in Fig.~\ref{fig:table}(14-29), power grasps provide multi-finger enclosure of cylindrical, spherical, and rod-like objects. Precision grasps include three-finger and fingertip pinches for small objects such as keys and wires, while tool-specific grasps demonstrate functional postures for scissors, screwdrivers, tape rolls, tweezers, and related tools.

\section{Discussion and Limitations}
The present evaluation focuses on quasi-static joint-angle prediction
under monotonic actuation. Reverse-direction effects, trial-to-trial
variability, and dynamic closed-loop tracking are not quantified, and
the grasping results are intended as qualitative functional
demonstrations rather than a statistical robustness benchmark.
Extending the framework to bidirectional and dynamic evaluation, as
well as direct comparison with alternative calibration methods, is
left for future work.

\section{Conclusion}

This paper presented H-PAC, a 6-actuator, 15-DoF underactuated tendon-driven hand with a control-oriented framework for repeatable posture generation. A sparse analytical actuator-joint model was derived from the tendon-routing geometry and combined with a tendon-elasticity compensation model to account for configuration-dependent deviations caused by restoring-spring loading. The method was implemented in a hierarchical architecture, where a host computer performs workspace-constrained mapping and compensation, while an ESP32 executes synchronized servo commands.

Monotonic servo-sweep experiments showed that the compensation improved joint-angle prediction over the tested ranges. The MAE of the index DIP joint decreased from $1.15^\circ$ to $0.18^\circ$, and all nine evaluated joints remained below $0.23^\circ$ after compensation. Representative postures and grasping configurations were also executed using the same model, control parameters, and embedded pipeline without task-specific retuning or external joint and force sensing in the control loop. These results demonstrate a practical approach to improving posture reproducibility in compact underactuated robotic end-effectors.

\bibliographystyle{IEEEtran} 
\bibliography{references}
\end{document}